\documentclass[review]{elsarticle}

\usepackage{lineno,hyperref}
\usepackage{graphicx}
\usepackage{multirow}
\graphicspath{ {./images/} }
\modulolinenumbers[5]

\begin{document}

\begin{frontmatter}

\title{ResAttUNet: Detecting Marine Debris using an Attention activated Residual UNet}

\author{Azhan Mohammed}
\address[]{GalaxEye Space}

\begin{abstract}
Currently, a significant amount of research has been done in field of Remote Sensing with the use of deep learning techniques. The introduction of Marine Debris Archive (MARIDA), an open-source dataset with benchmark results, for marine debris detection opened new pathways to use deep learning techniques for the task of debris detection and segmentation. This paper introduces a novel attention based segmentation technique that outperforms the existing state-of-the-art results introduced with MARIDA. The paper presents a novel spatial aware encoder and decoder architecture to maintain the contextual information and structure of sparse ground truth patches present in the images. The attained results are expected to pave the path for further research involving deep learning using remote sensing images. The code is available at \href{https://github.com/sheikhazhanmohammed/SADMA.git}{https://github.com/sheikhazhanmohammed/SADMA.git}

\end{abstract}


\end{frontmatter}

\section{Introduction}

Marine debris is a rising concern for numerous reasons that envelope environmental, economic, and human health aspects. Plastic waste in the ocean can float for a very long time, and has been found in many areas throughout the world \cite{S_Punla2022-nr}\cite{George2019-kk}. Numerous methods have been introduced to detect marine debris using the available open-source and commercial data sources\cite{Maximenko2019}, \cite{Hu2021}, \cite{ValdenegroToro2016}, like the use of satellite imagery\cite{Hu2021}, and autonomous underwater vehicles\cite{ValdenegroToro2016}.
Earth observation data from public and commercial satellite programs are very useful in detection of debris sites\cite{Topouzelis2019}, \cite{Kikaki2020}, \cite{Kremezi2021}, \cite{AcuaRuz2018}, \cite{Biermann2020}. To better understand the spectral behaviour of marine debris, indices like Floating Debris Index (FDI)\cite{Themistocleous2020} and Plastic Index (PI)\cite{Biermann2020} have been used.  Differentiating floating debris from other bright features like wave, sunlight, clouds, ships, foam, comes with challenges\cite{Garaba2018},\cite{Dierssen2020} that arise due to the fact that plastics have complex properties and vary in color, chemical composition, size and submergence level in water body \cite{Kikaki2020},\cite{Topouzelis2019}. 
Most of the earlier datasets focused on detecting objects like vessels\cite{Heiselberg2017}, clouds over sea area\cite{Kristollari2020} and macro algae\cite{Wang2021}. To overcome the existing challenges in the field MARIDA- MARIne Debris Archive\cite{10.1371/journal.pone.0262247} introduced an open-access benchmark dataset, that contains images extracted using Sentinel-2 multispectral satellite data. Along with the dataset, MARIDA introduced baseline results for semantic segmentation task using weakly supervised labels using Machine Learning algorithms and deep neural network architectures.
The contribution of this paper is a Marine Debris Detector using Sentinel S2 Satellite Imagery and a Attention Based Residual UNet, a novel convolution neural network specifically designed to segment sparse debris. The detection technique provides state-of-the-art results on the MARIDA open-source dataset that contains real world images that are temporally and geographically well-distributed.

\section{Methods}
\subsection{MARIDA Dataset}
MARIDA was constructed as a step by step process that included:
\begin{itemize}
    \item Collection of ground-truth data and relevant literature related to floating debris in coastal areas. The ground-truth data was collected for a period of 7 years, from 2015 to 2021, across coastal areas and river mouths in several countries.
    \item Acquiring satellite data, processing images, calculating spectral indices, annotating the collected images to formulate ground truth maps and performing statistical analysis on the generated ground truth database.
    \item Generating MARIDA Dataset and formulating ML and UNet benchmarks for weakly supervised image segmentation task.
\end{itemize}

\begin{table}[]
\centering
\caption{MARIDA class distribution}
\resizebox{\columnwidth}{!}{%
\begin{tabular}{|c|c|c|c|}
\hline
Class Name & Acronym & Total Pixels & Pixel Percentage \\ \hline
Marine Debris & MD & 3399 & 0.4059103606 \\ \hline
Dense Sargassum & DenS & 2797 & 0.3340192052 \\ \hline
Sparse Sargassum & Sps & 2357 & 0.2814741747 \\ \hline
Natural Organic Material & NatM & 864 & 0.1031793326 \\ \hline
Ship & Ship & 5803 & 0.6929972999 \\ \hline
Clouds & Cloud & 117400 & 14.0199695 \\ \hline
Marine Water & MWater & 129159 & 15.42423544 \\ \hline
Sediment-Laden Water & SLWater & 372937 & 44.5363319 \\ \hline
Foam & Foam & 1225 & 0.1462901417 \\ \hline
Turbid Water & TWater & 157612 & 18.82210761 \\ \hline
Shallow Water & SWater & 17369 & 2.074215079 \\ \hline
Waves & Waves & 5827 & 0.6958633925 \\ \hline
Cloud Shadows & CloudS & 11728 & 1.400563904 \\ \hline
Wakes & Wakes & 8490 & 1.013880247 \\ \hline
Mixed Water & MixWater & 410 & 0.04896241478 \\ \hline
\end{tabular}%
}
\label{tab:Table1}
\end{table}
MARIDA contains 1381 patches saved as geo-tif files along with their respective pixel-wise segmentation masks and confidence scores stored in a JSON file, collected from 63 scenes using the Sentinel S2 Satellite mission. Additionally, MARIDA also contains the shapefile data of these geo-tif files in WGS'84/UTM projection. The selected study sites are distributed over 11 countries, Honduras, Guatemala, Haiti, Santo Domingo, Vietnam, South Africa, Scotland, Indonesia, Philippines, South Korea and China.

The dataset has 15 different classes, that are tabulated in \autoref{tab:Table1}, the table also discusses the pixel level distribution of classes and indicates the under representation of various classes, a problem that is tackled using Weighted Cross Entropy loss, explained in detail in upcoming section.
A simpler approach to detect marine debris can be to transform the data into a binary classification problem, having just plastic and non-plastic classes, this makes the target problem easier and achieves better scores when compared to the baseline results introduced in MARIDA\cite{Booth2022HighprecisionDM}.

\subsection{UNet}
The UNet\cite{DBLP:journals/corr/RonnebergerFB15} network architecture has been used in various image segmentation tasks in various domains, including satellite imagery\cite{9760787},\cite{9426212},\cite{Chaudhary2022},\cite{Larionov2020}. The network consists of a contracting path and an expanding path. The contracting path is similar to an ordinary Convolution Neural Network consisting of Convolution layers, followed by a LeakyReLU activation and Max Pooling layers. After each downsample block, the image height and width are reduced by half while the feature channels are doubled. The expansive block consists of an upsampling layer that doubles the height and width of image using a bilinear upsampling, followed by convolution blocks to reduce the number of features. Before expanding each feature map, it is concatenated with the corresponding feature map from the contracting path. The concatenation of features from contracting path ensures transfer of spatial information from the contracting path to the expanding path. The final layer is a convolution layer with 1x1 kernel, it takes in 32 channels and outputs channels corresponding to number of classes.

\subsection{Residual Blocks}
Training of deep learning networks faced many issues, including training instability. Residual Blocks\cite{DBLP:journals/corr/HeZRS15} were introduced to help in stable training of very deep neural networks with the help of skip connections. It was noticed that when deep networks start converging, a degradation problem is exposed, where the accuracy first converges, then remains stables and then starts degrading. This degradation problem shows that deep models are not easy to optimize. To help with the issue, Residual blocks were introduced, where instead of hoping a few stacked layers to directly fit a desired output mapping \(H(x)\), the layers were made to fit simpler mapping \(F(x) = H(x)-x\), and with the help of skip connections, the mapping is then transformed into \(F(x)+x\) that was the original mapping. 

It has been noticed that using residual blocks in neural networks has a significant impact on the model's performance\cite{2019RoadTE},\cite{Wu2021HrlinknetLW},\cite{Sanya2019EXPLOITINGSC}. The introduced residual blocks in the model are employed after downsampling layers, and help in extracting deep features from the feature space obtained.

\subsection{Convolutional Block Attention Module}
\begin{figure}[t]
\caption{\textbf{Overview of CBAM.} The attention module has two sub-modules, Channel Attention Module and Spatial Attention Module}
\centering
\includegraphics[width=12cm]{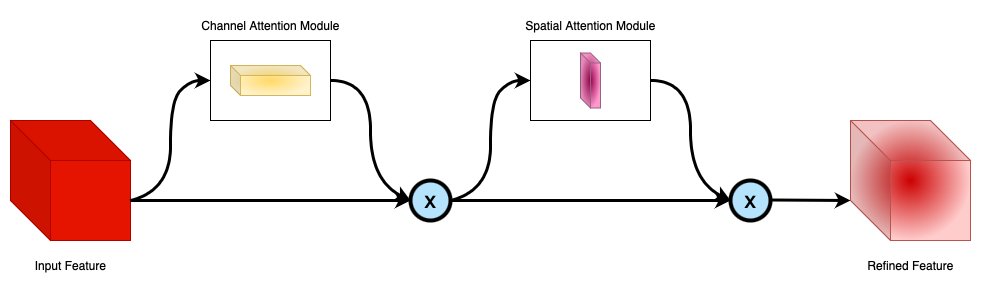}
\label{fig:CBAM}
\end{figure}

Convolutional Block Attention Module\cite{Woo2018}, hereby referred to as CBAM is a means to infer attention map in two separate dimensions. The module has two sequential sub-modules, channel attention module and spatial attention module. For any given input feature map \(F \in R^{C \times H \times W}\) the CBAM module's Channel Attention Module first generates a 1D attention map \(M_{c} \in R^{C \times 1 \times 1}\), followed by a 2D Attention Map \(M_{c} \in R^{1 \times H \times W}\) generated by Spatial Attention Module. The overall attention calculation can be written as:
\[ F' = M_{c}(F) \otimes F\]
\[ F'' = M_{s}(F') \otimes F'\]
Where \(\otimes\) refers to an element-wise multiplication. \autoref{fig:CBAM} shows the overall structure of CBAM.

\subsubsection{Channel Attention Module}
The channel attention map is generated by leveraging the inter-channel relationship of features. Given an input feature map, the module first calculates the maximum and average value for each channel. The extracted features are then passed through a multilayer perceptron network (MLP Network), followed by element wise sum. The resultant is then passed through a sigmoid layer to generate the resultant channel attention map. \autoref{fig:CAM and SAM} depicts the architecture of Channel Attention Module.

\subsubsection{Spatial Attention Module}
\begin{figure}[t]
\caption{Channel Attention Module and Spatial Attention Module}
\centering
\includegraphics[width=12cm]{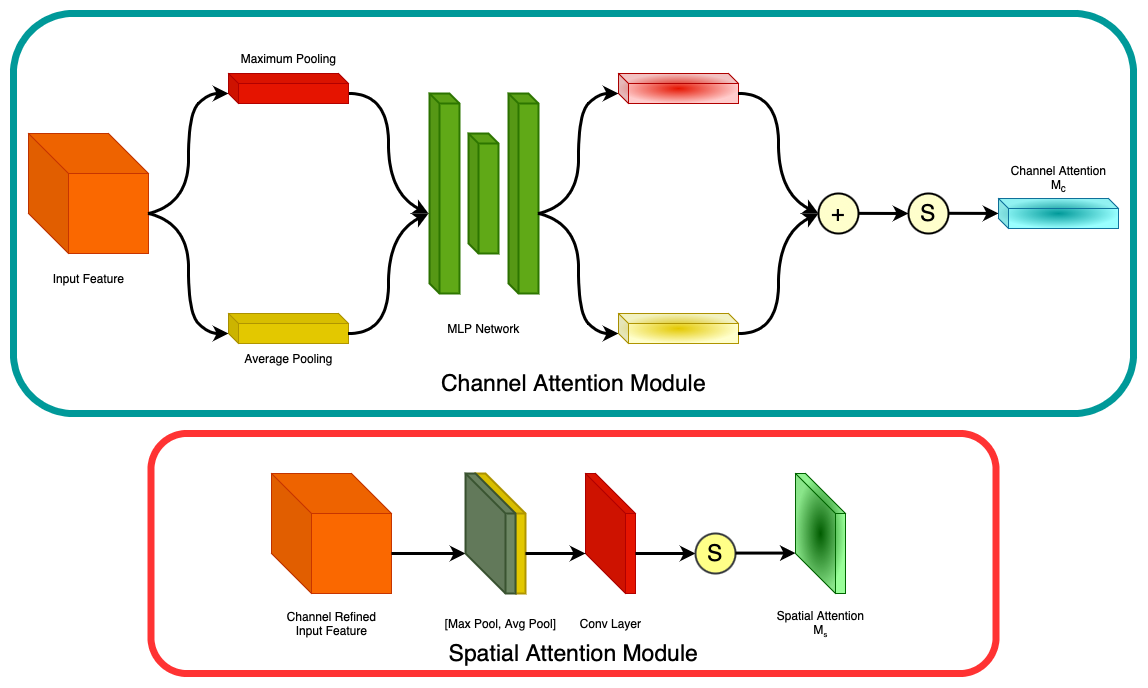}
\label{fig:CAM and SAM}
\end{figure}

Similar to Channel Attention Module, the Spatial Attention Module uses inter-spatial relationship of features to generate spatial attention map. For a channel refined input feature, the Spatial Attention Module first calculates maximum and average value across the channels, followed by a convolution layer and sigmoid layer to generate the spatial attention map. \autoref{fig:CAM and SAM} depicts the architecture of Spatial Attention Module.

\subsection{ResAttUNet}
\begin{figure}[h]
\caption{\textbf{ResAttUNet building blocks}}
\centering
\includegraphics[width=\textwidth]{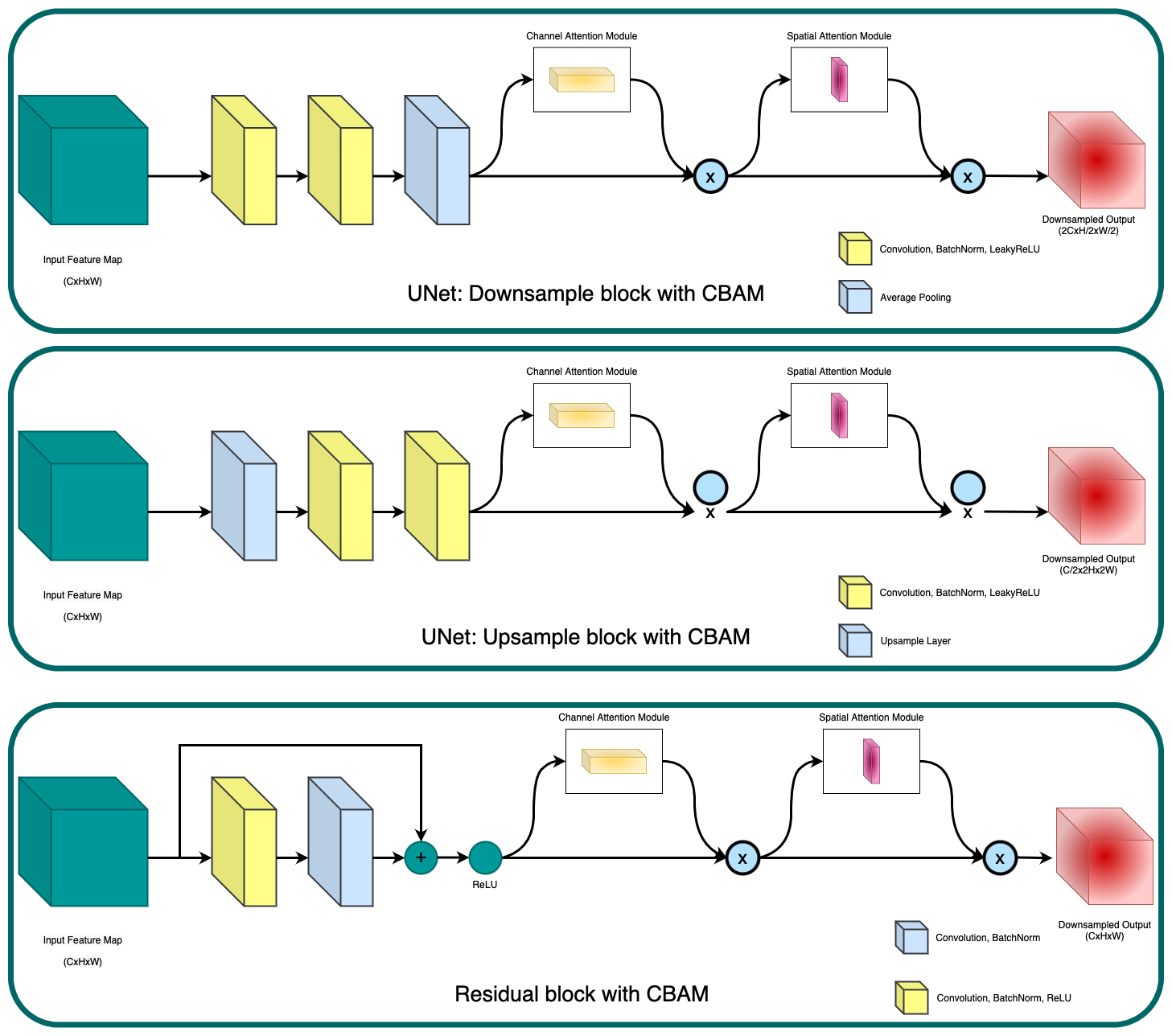}
\label{fig:res-attunet-blocks}
\end{figure}

\begin{figure}[h]
\caption{\textbf{ResAttUNet Model Architecture.} The downsample block with CBAM is shown in \autoref{fig:res-attunet-blocks}, upsample block with CBAM is shown in \autoref{fig:res-attunet-blocks} and the residual block with CBAM is shown in \autoref{fig:res-attunet-blocks}}
\centering
\includegraphics[width=\textwidth]{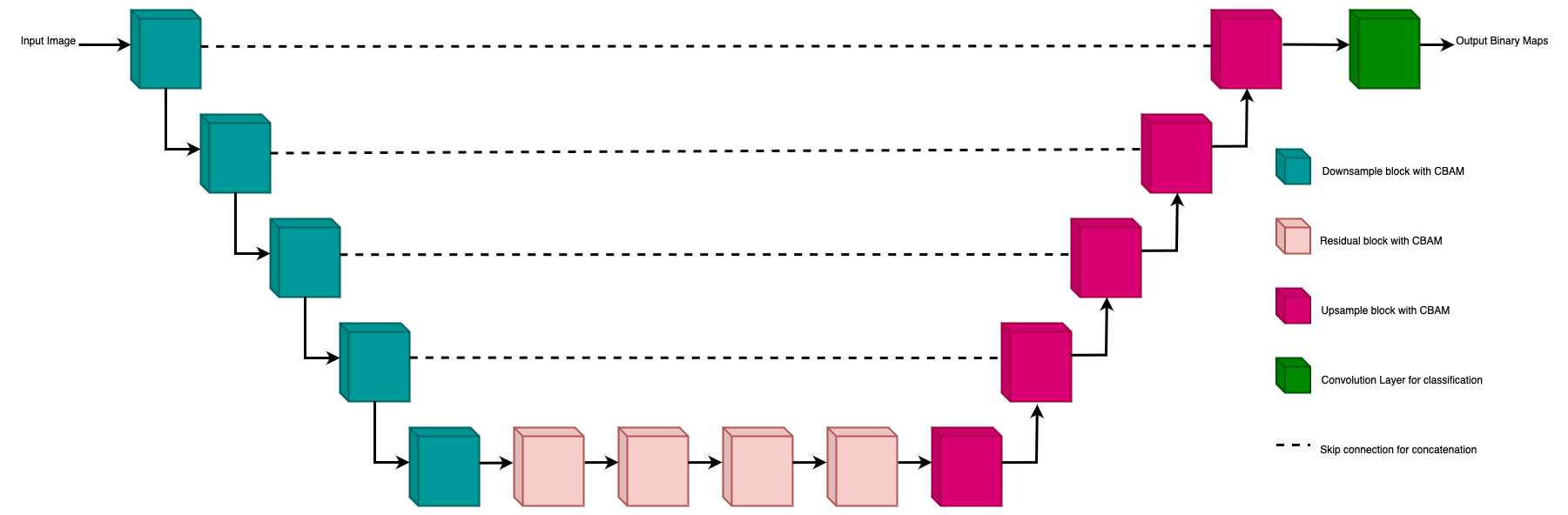}
\label{fig:ResAttUNet}
\end{figure}

The introduced novel architecture uses CBAM at every downsampling and upsampling stage of the UNet model. \autoref{fig:res-attunet-blocks} shows a CBAM activated downsampling layer. Similar to the downsampling block, the upsampling block uses CBAM as well, the architecture is depicted in \autoref{fig:res-attunet-blocks}. The final downsampling layer is followed by 3 residual blocks for deep feature extraction. A sample residual block is shown in \autoref{fig:res-attunet-blocks}. The complete network architecture is shown in \autoref{fig:ResAttUNet}.

\section{Methodology and Results}

Various experiments were conducted to finalize the final model architecture. A plain UNet model was first trained to achieve baseline results. The model downsample and upsample blocks were then incorporated with attention module. Followed by this, the residual layers were added, and later the residual layers were also incorporated with attention module to achieve the final model.

\begin{figure}[h]
\caption{Evaluation metrics for UNet-MARIDA, UNet, Attention UNet, and Residual Attention UNet plotted as a graph}
\centering
\includegraphics[width=\textwidth]{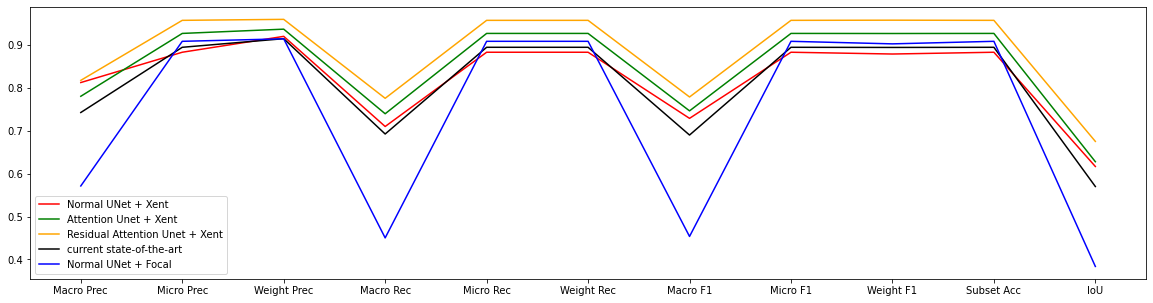}
\label{fig:modelcomparision}
\end{figure}

\begin{figure}[]
\caption{\textbf{Training loss vs Epoch} for UNet, Attention UNet and Residual Attention UNet}
\centering
\includegraphics[width=\textwidth]{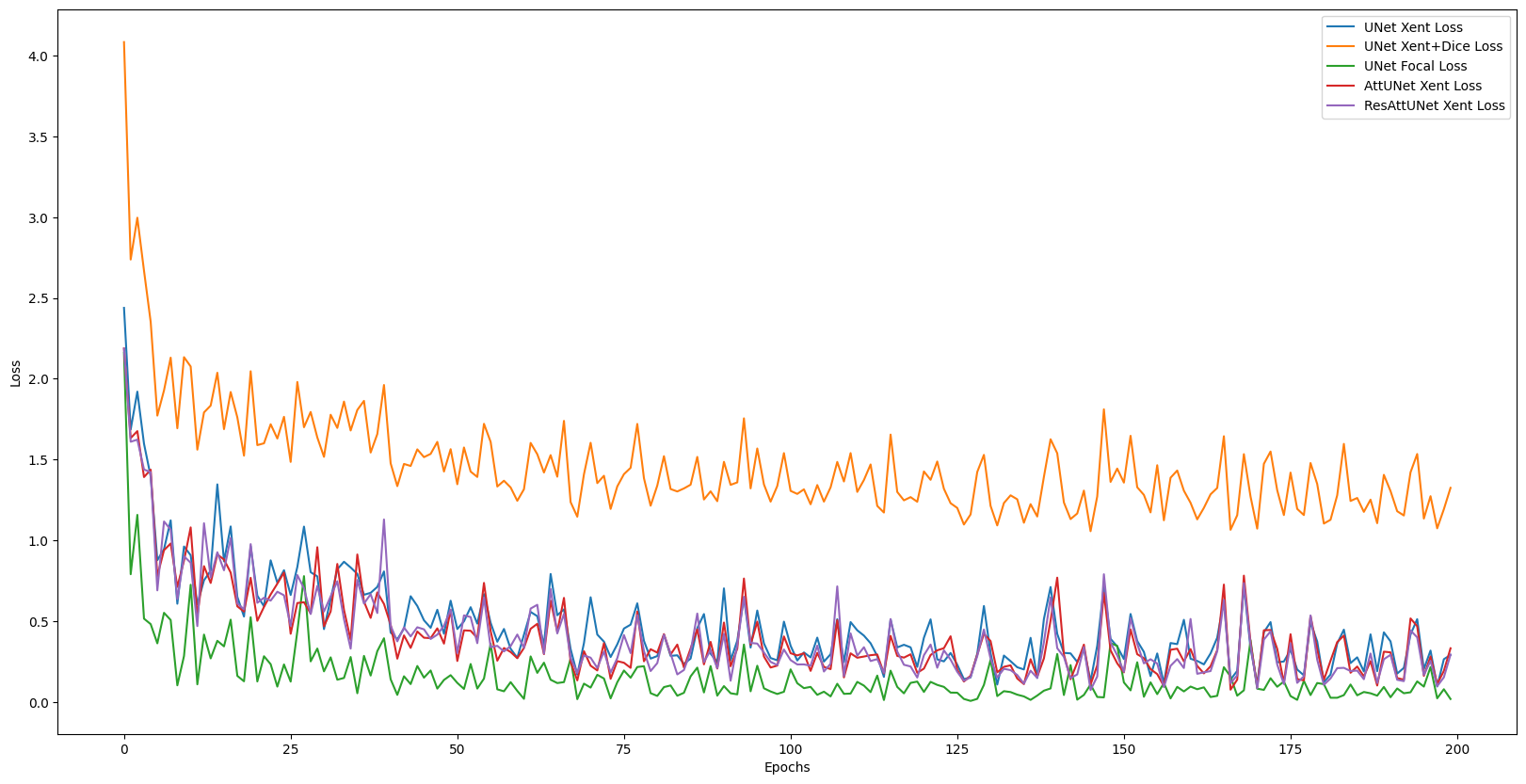}
\label{fig:modeltraininglosscomparision}
\end{figure}

\begin{figure}[]
\caption{\textbf{Validation loss vs Epoch} for UNet, Attention UNet and Residual Attention UNet}
\centering
\includegraphics[width=\textwidth]{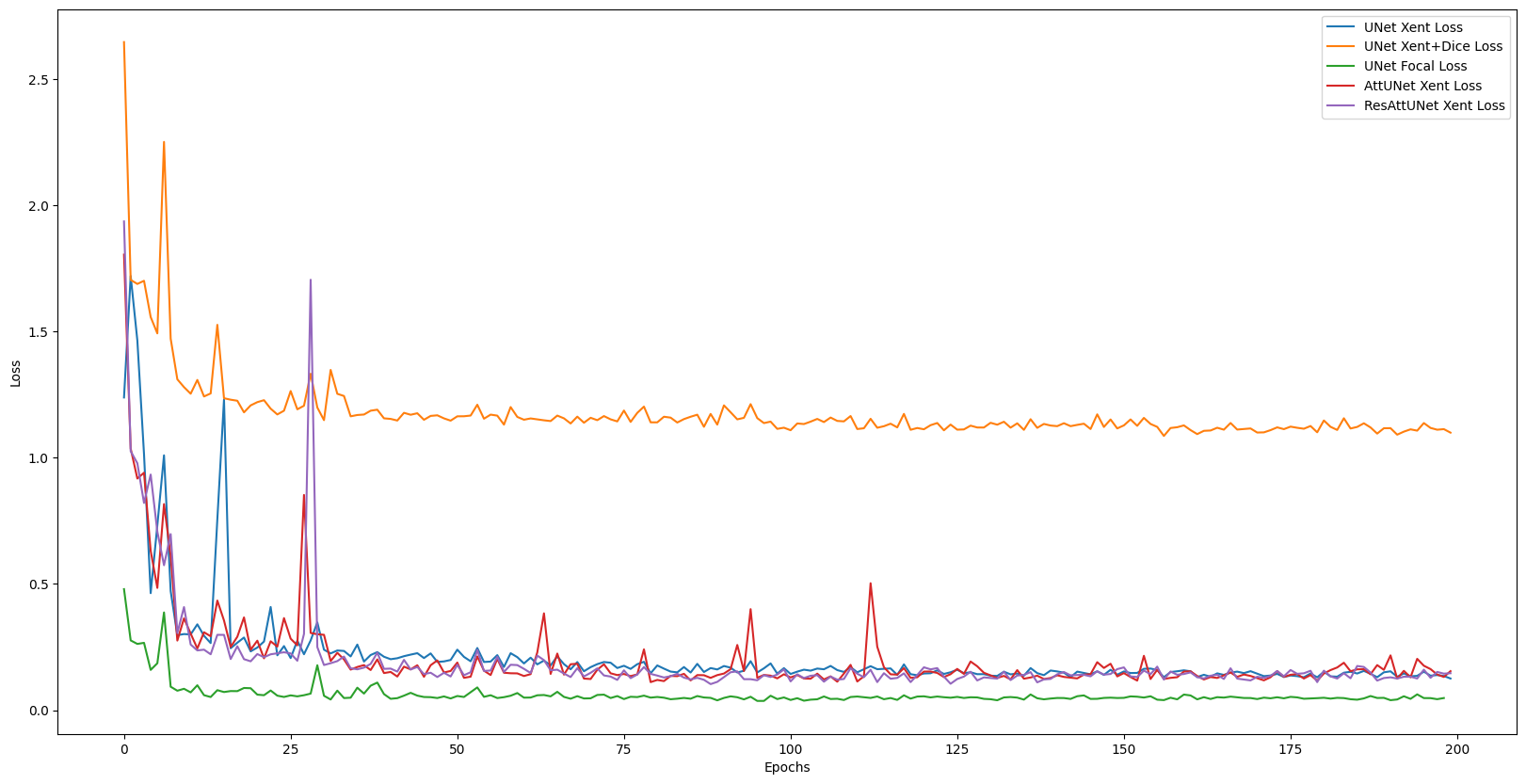}
\label{fig:validationlosscomparision}
\end{figure}

\begin{table}[]
\centering
\resizebox{\columnwidth}{!}{%
\begin{tabular}{|l|c|ccc|c|c|}
\hline
Model & \multicolumn{1}{l|}{UNet-MARIDA} & \multicolumn{3}{c|}{UNet} & Attention UNet & Residual Attention UNet \\ \hline
Loss Function & \multirow{2}{*}{Xent} & \multicolumn{1}{c|}{\multirow{2}{*}{Xent}} & \multicolumn{1}{c|}{\multirow{2}{*}{Focal}} & \multicolumn{1}{l|}{\multirow{2}{*}{Dice}} & \multirow{2}{*}{Xent} & \multirow{2}{*}{Xent} \\ \cline{1-1}
Metric &  & \multicolumn{1}{c|}{} & \multicolumn{1}{c|}{} & \multicolumn{1}{l|}{} &  &  \\ \hline
Macro Precision & 0.74 & \multicolumn{1}{c|}{0.81} & \multicolumn{1}{c|}{0.57} & 0.69 & 0.78 & \textbf{0.81} \\ \hline
Micro Precision & 0.89 & \multicolumn{1}{c|}{0.88} & \multicolumn{1}{c|}{0.90} & 0.90 & 0.92 & \textbf{0.95} \\ \hline
Weight Precision & 0.91 & \multicolumn{1}{c|}{0.92} & \multicolumn{1}{c|}{0.91} & 0.92 & 0.93 & \textbf{0.96} \\ \hline
Macro Recall & 0.69 & \multicolumn{1}{c|}{0.71} & \multicolumn{1}{c|}{0.45} & 0.61 & 0.74 & \textbf{0.77} \\ \hline
Micro Recall & 0.89 & \multicolumn{1}{c|}{0.88} & \multicolumn{1}{c|}{0.90} & 0.90 & 0.92 & \textbf{0.95} \\ \hline
Weight Recall & 0.89 & \multicolumn{1}{c|}{0.88} & \multicolumn{1}{c|}{0.90} & 0.90 & 0.92 & \textbf{0.95} \\ \hline
Macro F1 & 0.69 & \multicolumn{1}{c|}{0.72} & \multicolumn{1}{c|}{0.45} & 0.63 & 0.74 & \textbf{0.77} \\ \hline
Micro F1 & 0.89 & \multicolumn{1}{c|}{0.88} & \multicolumn{1}{c|}{0.90} & 0.90 & 0.92 & \textbf{0.95} \\ \hline
Weight F1 & 0.89 & \multicolumn{1}{c|}{0.87} & \multicolumn{1}{c|}{0.90} & 0.90 & 0.92 & \textbf{0.95} \\ \hline
Subset Accuracy & 0.89 & \multicolumn{1}{c|}{0.88} & \multicolumn{1}{c|}{0.90} & 0.90 & 0.92 & \textbf{0.95} \\ \hline
IoU & 0.57 & \multicolumn{1}{c|}{0.61} & \multicolumn{1}{c|}{0.38} & 0.54 & 0.62 & \textbf{0.67} \\ \hline
\end{tabular}%
}
\caption{Evaluation scores obtained by: UNet-MARIDA, UNet, Attention UNet, and Residual Attention UNet. The loss function used to train the model is mentioned in the sub-column under the model. The best score for each metric is highlighted in bold text. A graphical representation for the metrics can be found in \autoref{fig:modelcomparision}}
\label{tab:evaluation-table}
\end{table}

As discussed earlier, the MARIDA dataset has a class imbalance, to deal with imbalanced classes, the model was trained using weighted cross entropy loss. The weights are derived using log of class frequency in the dataset. The weighted cross-entropy loss is given by:
\[xentloss_{weighted} = -\frac{1}{N}\sum_{n=1}^{N}\omega r_n log(p_n) + (1-r_n)log(1-p_n)\]
Where, 
\[\omega=\frac{N-\Sigma_n p_n}{\Sigma_n p_n}\]
Experiments were conducted using a sum of Weighted Cross Entropy Loss and Dice Loss to help with class imbalance, but dice loss only focuses on imbalance problem between the foreground and background, and overlooks the imbalance between hard samples and easy samples\cite{9338261}, and the same result was observed in our experiment. The trained model failed to perform at par with the baseline model.

Experiments were also conducted using Focal loss\cite{8417976} to deal with the class imbalance in the dataset. Focal loss is a modified version of existing cross entropy loss function, that attaches a modulating term that focuses on learning the hard misclassified samples. The focal loss function is given as:
\[focal loss = -(1-p_t)^\gamma log(p_t)\]
Where \(p_t\) is the predicted probability, and \(\gamma\) is the tuning control parameter. For our experiments, value of \(\gamma\) was set as 2. As mentioned earlier, focal loss is a modified cross entropy loss function, so we did not use weighted cross entropy loss with focal loss to train the model. The validation loss for the model trained using focal loss converges pretty well as seen in \autoref{fig:validationlosscomparision}, but the evaluation metrics reveal that the model was overfitting when tested on the test data. Hence focal loss was also discarded while training the final model.
The finalized network architecture was trained using Weighted Cross Entropy loss, for 200 epochs with an initial learning rate of \(1e-3\) and the learning rate was gradually reduced by a factor of \(0.5\) at interval of 40 epochs. The final values of Precision, Recall, F1 score and IoU for each model is mentioned in \autoref{tab:evaluation-table}.
As seen in validation loss trend, shown in \autoref{fig:validationlosscomparision}, the Residual Attention UNet model outperforms the other models by a fair margin. The evaluation metrics shown in \autoref{fig:modelcomparision} also shows the superiority of ResAttUNet model over other architectures and training procedures. This shows that using residual blocks with attention has great potential in detection of sparse debris scattered throughout the sea, and can be used for various applications.

\section{Conclusion}

Debris detection using Satellite imagery can be very helpful in tackling the problems caused by presence of debris in the sea and oceans. In this study, we propose a neural network capable of segmenting not only accumulated debris and sediment, but also segment out sparse debris particles from satellite imagery with increased accuracy and lesser rate of false positives. The overall F1 score achieved is 0.95, with an IoU score of 0.65. Proposed methodology is lightweight can be easily incorporated to perform debris segmentation using multispectral satellite imagery. In the upcoming future work, Synthetic aperture radar (SAR) imagery can be used along with Multi-spectral images to increase the performance of models and derive more useful information from the images.

\bibliography{mybibfile}

\end{document}